\documentclass{article}
\usepackage{proceed2e}
\usepackage{graphicx}
\usepackage{subfigure}
\usepackage{amsmath,mathrsfs,amssymb}
\title{  Percolation Thresholds of Updated Posteriors\\ for Tracking Causal Markov Processes in Complex Networks} 

\author{ 
{\bf Patrick L. Harrington Jr. \thanks{ \ \ \texttt{plhjr@umich.edu} }} \\  
Bioinformatics Graduate Program \\
Department of Statistics\\  
University of Michigan\\ 
Ann Arbor, MI 48109 \\ 
\And 
{\bf Alfred O. Hero III. \thanks{ \ \ \texttt{hero@umich.edu}}}  \\ 
Department of EECS          \\ 
Department of Statistics\\
University of Michigan \\   
Ann Arbor, MI, 48109           
} 
 
\begin{document} 
 
\maketitle

\begin{abstract} 
Percolation on complex networks has been used to study computer viruses, epidemics, and other casual processes.  Here, we present conditions for the existence of a network specific, observation dependent, phase transition in the updated posterior of node states resulting from actively monitoring the network.  Since traditional percolation thresholds are derived using observation independent Markov chains, the threshold of the posterior should more accurately model the true phase transition of a network, as the updated posterior more accurately tracks the process.  These conditions should provide insight into modeling the dynamic response of the updated posterior to active intervention and control policies while monitoring large complex networks.

\end{abstract} 
 
\section{INTRODUCTION}  
Increasingly often, researchers are confronted with monitoring the states of nodes in large computer, social, or power networks where these states dynamically change due to viruses, rumors, or failures that propagate according to the graph topology \cite{cohen-2003-91,Eubank_2004,newman-2002-66}.  This class of network dynamics has been extensively modeled as a percolation phenomenon, where nodes on a graph can randomly ``infect'' their neighbors.

Percolation across networks has a rich history in the field of statistical physics, computer science, and mathematical epidemiology. Here, researchers are typically confronted with a network, or a distribution over the network topology, and extract fixed point attractors of node configurations, thresholds for phase transitions in node states, or distributions of node state configurations \cite{draief-2006,chak08,newman-2005-95}.  In the field of fault detection, the nodes or edges can ``fail'', and the goal is to activate a subset of sensors in the network which yield high quality measurements that identify these failures  \cite{Zheng05}.  While the former field of research concerns itself with extracting \textit{offline} statistics about properties of the percolation phenomenon on networks, devoid of any measurements, the latter field addresses \textit{online} measurement selection tasks.

Here, we propose a methodology that actively tracks a causal Markov process across a complex network (such as the one in Figure \ref{sf}), where measurements are adaptively selected. We extract conditions such that the updated posterior probability of all nodes ``infected'' is driven to one in the limit of large observation time.  In other words, we derive conditions for the existence of an epidemic threshold on the updated posterior distribution over the states.  

The proposed percolation threshold should more accurately reflect the true conditions that cause a phase transition in a network, e.g., node status changing from healthy/normal to infected/failed, than traditional thresholds derived from conditions on predictive distributions which are devoid of observations or controls.      

Since most practical networks of interest are large, such as electrical grids, it is usually infeasible to sample all nodes continuously, as such measurements are either expensive or bandwidth is limited.  Given these, or other resource constraints, we present an information theoretic sampling strategy that selectively targets specific nodes that will yield the largest information gain, and thus, better detection performance.  

The proposed sampling strategy balances the trade-off between trusting the \textit{predictions} from the known model dynamics (from percolation theory) and expending precious resources to select a set of nodes for measurement.     


We present the adaptive measurement selection problem and give two tractable approximations to this subset selection problem based upon the joint and marginal posterior distribution, respectively.  A set of decomposable Bayesian filtering equations are presented for this adaptive sampling framework and the necessary tractable inference algorithms for complex networks are discussed.  We present analytical worst case performance bounds for our adaptive sampling performance, which can serve as sampling heuristics for the activation of sensors or trusting predictions generated from previous measurements.

To the author's knowledge, this is the first attempt to extract a percolation threshold of an actively monitored network using the updated posterior distribution instead of the observation independent predictive distributions.

\section{PROBLEM FORMULATION} 
The objective of actively monitoring the $n$ node network is to recursively update the posterior distribution of each hidden node state given various measurements.   Specifically, the next set of $m$ measurement actions (nodes to sample), $m \ll n$, at next discrete time are chosen such that they yield the highest quality of \textit{information} about the $n$ hidden states.  The condition on $m\ll n$ simulates the reality of fixed resource constraints, where typically only a small subset of nodes in a large network can be observed at any one time.

\begin{figure}[h!]
\centering
\includegraphics[width=1.75in]{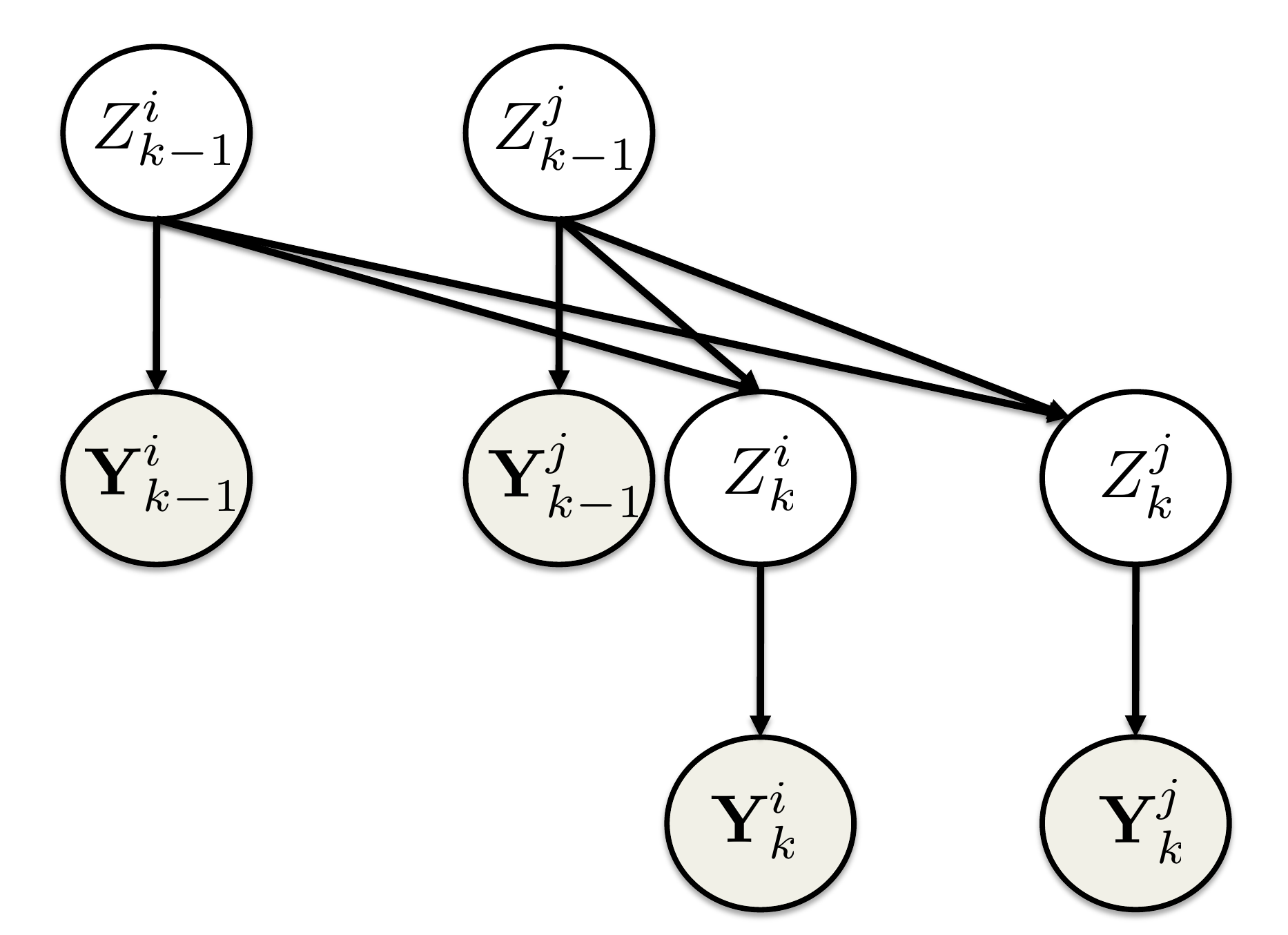}
\caption{Partially Observed Markov Structure for $i$ and $j$ for $\mathcal{E}\left(\mathcal{V}_i,\mathcal{V}_j\right) \notin \emptyset$}
\label{2tbn}
\end{figure}

Here, the hidden states are discrete random variables that correspond to the states encoded by the percolation process on the graph.  Here, the graph $\mathcal{G} = (\mathcal{V},\mathcal{E})$, with $\mathcal{V}$ representing the set of nodes and $\mathcal{E}$ corresponding to the set of edges.  Formally, we will assume a state-space representation of a discrete time, finite state, partially observed Markov decision process (POMDP). Here,
\begin{equation}\label{Z}
	\textbf{Z}_k = \{Z_k^1,\dots,Z_k^n\}
\end{equation}
represents the joint hidden states, e.g., healthy or infected
\begin{equation}\label{Y} 
	\textbf{Y}_k = \{\textbf{Y}_k^{(1)},\dots,\textbf{Y}_k^{(m)}\}
\end{equation} 
represents the $m$ observed measurements obtained at time $k$, e.g., biological assays or {\tt PING}ing an IP address, and 
\begin{equation}\label{A}
	\textbf{a}_k = \{a^1_k,\dots,a^m_k\}
\end{equation} 
represents the $m$ actions taken at time $k$, i.e., which nodes to sample.  Here, $\textbf{Y}_k^{(j)}$, continuous/categorical valued vector of measurements, which is induced by action $a^j_k$, $a^j_k \in \mathcal{A}$, with $\mathcal{A} = \{1,\dots,n\}$ confined to be the set of all $n$ individuals in the graph, and $Z_k^i \in \{0,1, \dots, r\}$.  Since the topology of $\mathcal{G}$ encodes the direction of "flow" for the process, the state equations may be modeled as a decomposable partially observed Markov process:   
\begin{eqnarray}\label{state_eqs}
	\textbf{Y}^i_k & =& \textbf{f}( Z^i_k ) + \textbf{w}^i_k\label{Y} \\
	Z^i_k & =& h\left( Z^i_{k-1}, \{ Z^j_{k-1} \}_{j \in \eta(i) } \right)\label{Z}.
\end{eqnarray}  
Here, $\eta(i) = \{j: \mathcal{E}\left(\mathcal{V}_i,\mathcal{V}_j \right) \notin \emptyset \}$ is the neighborhood of $i$, $\textbf{f}( Z^i_k )$ is a non-random vector-valued function, $\textbf{w}^i_k$ is measurement noise, and $h\left( Z^i_{k-1}, \{ Z^j_{k-1} \}_{j \in \eta(i) } \right)$ is a stochastic equation encoding the transition dynamics of the Markov process (see Figure \ref{2tbn} for a two node graphical model representation).   

\subsection{BAYESIAN FILTERING} 
In our proposed framework for actively monitoring the hidden node states in the network, the posterior distribution is the sufficient statistic for inferring these states.  The general recursion for updating the joint posterior probability given all past and present observations is given by the standard Bayes update formula:  
\begin{equation}\label{bayes}
	p(\textbf{Z}_k | \textbf{Y}_0^k) = \frac{ f( \textbf{Y}_k | \textbf{Z}_k ) }{ g( \textbf{Y}_k | \textbf{Y}_0^{k-1} ) } p( \textbf{Z}_k | \textbf{Y}^{k-1}_0 )
\end{equation}
with
\begin{equation}\label{intract}
	p( \textbf{Z}_k | \textbf{Y}^{k-1}_0 ) = \hspace{-6mm} \sum_{ \textbf{z} \in \{0,1,\dots,r\}^n } \hspace{-4mm} p( \textbf{Z}_k | \textbf{Z}_{k-1} = \textbf{z}) p( \textbf{Z}_{k-1} = \textbf{z} | \textbf{Y}^{k-1}_0).
\end{equation}
The Chapman-Kolmogorov equations provide the connection between the posterior update (\ref{intract}) and the distribution resulting from the standard percolation equations.  In the former, the updates are conditional probabilities that are conditional on past observations, while in the latter, the updates are not dependent on observations.  


The local interactions in the graph $\mathcal{G}$ imply the following conditional independence assumptions:  
\begin{equation}\label{likelihood}
	f( \textbf{Y}_k | \textbf{Z}_k) = \prod_{i=1}^n f( \textbf{Y}^i_k | Z^i_k).
\end{equation}
\begin{equation}\label{trans}
	p( \textbf{Z}_k | \textbf{Z}_{k-1}) = \prod_{i=1}^n p\left( Z^i_k | Z^i_{k-1}, \{  Z^j_{k-1}   \}_{j \in \eta(i)}\right)
\end{equation}
where the likelihood term is defined in (\ref{Y}) and the transition dynamics are defined in (\ref{Z}).  This decomposable structure allows the belief state (posterior excluding time $k$ observations) update, for the $i^{th}$ node in $\mathcal{G}$, to be written as:
\begin{equation}\label{marginal}
	p(Z_k|\textbf{Y}_0^{k-1}) =  \hspace{-7mm}   \sum_{\textbf{z} \in \{0,1,\dots,r\}^{\lVert \text{pa} \rVert}} \hspace{-7mm} p(Z_k | \textbf{Z}^{ \text{pa}  }_{k-1} = \textbf{z})p(\textbf{Z}^{  \text{pa}     }_{k-1} = \textbf{z} |\textbf{Y}_0^{k-1})
\end{equation}
with the parent set, $\text{pa} = \{ \eta(i), i\}$.  Unfortunately, for highly connected nodes in $\mathcal{G}$, this marginal update becomes intractable.  It thus must be approximated \cite{Doucet:2000bv, Ng02factoredparticles, Mackay_2002_information}.

\subsection{INFORMATION THEORETIC ADAPTIVE SAMPLING}
In most real world situations, acquiring measurements from all $n$ nodes at any time $k$ is unrealistic, and thus, a sampling policy must be exploited for measuring a subset of nodes \cite{hero_book_07,Zheng05}.  Since we are concerned with monitoring the states of the nodes in the network, an appropriate reward is the expected information gain between the \textit{updated} posterior, $p_k = p(\textbf{Z}_k | \{ \textbf{Y}^{i}_k\}_{i \in \textbf{a}_k}, \textbf{Y}_0^{k-1})$, and the belief state, $p_{k|k-1} = p(\textbf{Z}_k | \textbf{Y}_0^{k-1})$:
\begin{equation}\label{IG}
	\textbf{a}_k = \mbox{argmax}_{\textbf{a} \subset \mathcal{A}  }\mathbb{E}\left[ \mathcal{D}_{\alpha}\left( \{ \textbf{Y}^{i}_k \}_{i \in \textbf{a}} \right) | \textbf{Y}^{k-1}_0 \right]
\end{equation}	
\begin{equation}\label{div}
	\mathcal{D}_{\alpha}\left( \{ \textbf{Y}^{i}_k \}_{i \in \textbf{a}} \right) = \mathcal{D}_{\alpha}\left( p_k || p_{k|k-1}    \right), \ 0 < \alpha < 1
\end{equation}	
with $\alpha$-Divergence
\begin{equation}\label{alpha_div}
 	\mathcal{D}_{\alpha}(p || q) = \frac{1}{\alpha-1} \mbox{log} \left( \mathbb{E}_q \left[  \left(p/q\right)^{\alpha}   \right]   \right)
\end{equation}
for distributions $p$ and $q$ with identical support.

The reward  in (\ref{IG}) has been widely applied to multi-target, multi-sensor tracking for many problems including, sensor management and surveillance \cite{hero_book_07, ecs16801}.  Note that $\lim_{\alpha \rightarrow 1}\mathcal{D}_{\alpha}( p || q ) \rightarrow \mathcal{D}_{KL}( p || q)$, where $\mathcal{D}_{KL}( p ||q)$ is the Kullback-Leibler divergence between $p$ and $q$.  The expectation in (\ref{IG}) is taken with respect to the conditional distribution $g(\textbf{Y}_k|\textbf{Y}_0^{k-1})$ given the previous measurements $\textbf{Y}_0^{k-1}$ and actions $\textbf{a}_k$.  In practice, the expected information divergence in (\ref{IG}) must be evaluated via Monte-Carlo methods.  Also, the maximization in (\ref{IG}) requires enumeration over all $\binom{n}{m}$ actions (for subsets of size $m$), and therefore, we must resort to Greedy approximations.  We propose incrementally constructing the set of actions at time $k$, $\textbf{a}_k$, for $j = 1,\dots,m$, according to:
\begin{equation}\label{approxIG}
	a^j_k = \mbox{argmax}_{i \in \mathcal{A} \setminus \textbf{a}_k  }\mathbb{E}\left[ \mathcal{D}_{\alpha}\left( \textbf{Y}^i_k, \{ \textbf{Y}^{j}_k \}_{j \in \textbf{a}_k} \right) | \textbf{Y}^{k-1}_0 \right].
\end{equation}	
Both (\ref{IG}) and (\ref{approxIG}) are selecting the nodes to sample which yield maximal divergence between the percolation prediction distribution (belief state) and the updated posterior distribution, averaged over all possible observations.  Thus (\ref{IG}) provides a metric to assess whether to trust the predictor and defer actions until a future time or choose to take action, sample a node, and update the posterior.

\subsubsection{Lower Bound on Expected $\alpha$-Divergence}
Since the expected $\alpha$-Divergence in (\ref{IG}) is not closed form, we could resort to numerical methods for estimating this quantity.  Alternatively, one could specify an analytical lower-bound that could be used in-lieu of numerically computing the expected information gain in (\ref{IG}) or (\ref{approxIG}).  

We begin by noting that the expected divergence between the updated posterior and the predictive distribution (conditioned on previous observations) differ only through the measurement update factor, $f_k/g_{k|k-1}$ ((\ref{IG}) re-written):

\vspace{-7mm}
\begin{gather}
	\mathbb{E}_{g_{k|k-1}} \left[ \mathcal{D}_{\alpha} \left( p_k || p_{k|k-1} \right) \right]] \nonumber\\  = \mathbb{E}_{g_{k|k-1}} \left[ \frac{1}{\alpha-1} \mbox{log} \ \mathbb{E}_{p_{k|k-1}}  \left[ \left( \frac{ f_k }{g_{k|k-1}} \right)^{\alpha}  \right]    \right]\label{alphaDiv}.
\end{gather}
\vspace{-4mm}


So, if there is significant overlap between the likelihood distributions of the observations, the expected divergence will tend to zero, implying that there is not much value-added in taking measurements, and thus, it is sufficient to use the percolation predictions for inferring the states.  

It would be convenient to interchange the order of the conditional expectations in (\ref{alphaDiv}). It is easily seen that Jensen's inequality yields the following lower bound for the expected information gain
\begin{gather}\label{performbound}
	\mathbb{E}_{g_{k|k-1}} \left[ \mathcal{D}_{\alpha} \left( p_k || p_{k|k-1} \right) \right] \nonumber\\  \ge \frac{1}{\alpha-1} \mbox{log} \ \mathbb{E}_{p_{k|k-1}} \left[  \mathbb{E}_{g_{k|k-1}} \left[     \left( \frac{ f_k }{g_{k|k-1}} \right)^{\alpha}        \right]  \right].
\end{gather}
Here, the inner conditional expectation can be obtained from $\mathcal{D}_{\alpha}\left(f_k||g_{k|k-1} \right)$, which has a closed form for common distributions (e.g., multivariate Gaussians) \cite{hero_book_07}.

\section{ASYMPTOTIC ANALYSIS OF MARGINAL POSTERIOR}
For tracking the percolation process across $\mathcal{G}$, we have discussed recursive updating of the belief state.  However, computing these updates exactly is in general intractable.  For the remainder of the paper, we will use  (\ref{Y}) and (\ref{Z}) to directly update the marginal posterior distribution using the following matrix representation:
\begin{equation}\label{marg_update}
	\textbf{p}_k(z) = \textbf{D}_k(z) \textbf{p}_{k|k-1}(z)
\end{equation}
with updated marginal posterior $\textbf{p}_k(z) = [p_{1,k}(z), \dots,p_{n,k}(z)]^T$ with $p_{i,k}(z) = p(Z^i_k=z| \textbf{Y}^i_k, \textbf{Y}^{k-1}_0)$, $\textbf{D}_k(z) = \text{diag} \left(  f_{i,k}^{(z)} / g_{i,k|k-1} \right) $, and marginal belief state $\textbf{p}_{k|k-1}(z) = [p_{1,k|k-1}(z), \dots,p_{n,k|k-1}(z)]^T$ with $p_{i,k|k-1}(z) = p(Z^i_k=z | \textbf{Y}^{k-1}_0)$.  

Note that for $i \notin \textbf{a}_k$, $\left( \textbf{D}_k(z) \right)_{i,i} = 1$, and $p_{i,k}(z) = p_{i,k|k-1}(z)$.  Given that we can find an efficient way of updating $\textbf{p}_{k|k-1}(z)$, according to the transition dynamics (\ref{Z}), we can solve a modified version of (\ref{approxIG}), for $j=1,\dots,m$:
\begin{equation}\label{IG_marg}
	a^j_k = \mbox{argmax}_{i \in \mathcal{A} \setminus \textbf{a}_k }\mathbb{E}\left[ \mathcal{D}_{\alpha}\left( \textbf{Y}^i_k \right) | \textbf{Y}^{k-1}_0 \right]
\end{equation}  
\begin{equation}\label{div}
	\mathcal{D}_{\alpha}\left( \textbf{Y}^i_k \right) = \mathcal{D}_{\alpha}\left( p_{i,k}(z) || p_{i,k|k-1}(z)    \right), \ 0 < \alpha < 1. 
\end{equation}	 

\subsection{TOTAL DIVERGENCE OF UPDATED POSTERIOR} 
One interesting property of the Bayesian filtering equations is that the updated posterior can be written as a perturbation of the predictive percolation distribution through the following relationship ($z$ omitted for clarity):
\begin{equation}\label{mat_update}
	\textbf{p}_k = \textbf{D}_k \textbf{p}_{k|k-1} = \textbf{p}_{k|k-1} + \left( \textbf{D}_k - \textbf{I} \right) \textbf{p}_{k|k-1}. 
\end{equation}	
Hence, when the sensors do a poor job in discriminating the observations, $\textbf{D}_k \approx \textbf{I}$, we have $\textbf{p}_k \approx \textbf{p}_{k|k-1}$.  It is of interest to determine when there is significant difference between the posterior update and the prior update specified by the standard percolation equations.  Recall that the updated posterior is, in the mean, equal to the predictive distribution, $\mathbb{E}\left[ \textbf{p}_k |\textbf{Y}^{k-1}_0 \right] = \textbf{p}_{k|k-1}$.  The total deviation of the updated posterior from the percolation distribution can be summarized by computing the trace of the following conditional covariance:

\vspace{-5mm}
\begin{small}
\begin{gather}
	\mbox{tr} \left( \textbf{R} \left[ \textbf{p}_k | \textbf{Y}^{k-1}_0 \right] \right)  = \label{cond_cov} \\
	 \mbox{tr} \left( \mathbb{E} \left[ \left( \textbf{p}_k - \mathbb{E} \left[ \textbf{p}_k | \textbf{Y}^{k-1}_0 \right] \right) \left( \textbf{p}_k - \mathbb{E} \left[ \textbf{p}_k | \textbf{Y}^{k-1}_0 \right] \right)^T      | \textbf{Y}^{k-1}_0 \right] \right). \nonumber
\end{gather} 
\end{small}
\vspace{-5mm}

Using (\ref{mat_update}) and properties of the trace operator, we obtain the following measure of total deviation of the updated posterior from the predictive distribution in terms of $f_k$ and $g_{k|k-1}$:

\vspace{-5mm}
\begin{gather}\label{deviation}
	 \text{tr} \left( \textbf{R} \left[ \textbf{p}_k | \textbf{Y}^{k-1}_0 \right] \right) = \mbox{tr} \left(  \mathbb{E} \left[ \left( \textbf{D}_k - \textbf{I}\right)^2  | \textbf{Y}^{k-1}_0 \right] \textbf{P}_{k|k-1} \right)
\end{gather}
\vspace{-5mm}

with $\textbf{P}_{k|k-1} = \textbf{p}_{k|k-1}\textbf{p}_{k|k-1}^T$.  The conditional expectation in (\ref{deviation}) is the Pearson $\chi^2$ divergence between distributions $f_{i,k}$ and $g_{i,k|k-1}$, for all $i$.  This joint measure of deviation is analytical for particular families of distributions and thus can be used as an alternative measure of divergence for activation of sensors \cite{hero_book_07}.     

\subsection{PERCOLATION THRESHOLD OF UPDATED POSTERIOR}
There has recently been significant interest in deriving the conditions of a percolation/epidemic threshold in terms of transition parameters and the graph adjacency matrix spectra for two state causal Markov processes \cite{chak08,draief-2006,newman-2005-95}.  Such thresholds yield conditions necessary for an epidemic to arise from a small number of ``infections''.  Knowledge of these conditions are particularly useful for designing ``robust'' networks, where the probability of epidemics is minimized.    

Percolation thresholds are typically obtained by extracting the sufficient conditions of the network and model parameters for the node states to be driven to their stationary point, with high probability.  The probability of these events are usually computed using the observation independent percolation distribution  \cite{chak08,draief-2006,newman-2005-95}.

We use the results in \cite{chak08,draief-2006} to derive a percolation threshold based upon the updated posterior distribution assuming a restricted class of two-state Markov processes.  These conditions should more accurately model the \textit{current} network threshold since the posterior distribution tracks a particular ``disease'' trajectory better than the observation independent percolation distribution.       

Formally, $Z^i_k \in \{0,1\}$, $f^{(z)}_{i,k} = f(\textbf{Y}^i_k | Z^i_k = z)$ is the conditional likelihood for node $i$, $p_{i,k} = p(Z^i_k = 1 | \textbf{Y}^i_k,\textbf{Y}^{k-1}_0)$, and $p_{i,k} = p(Z^i_k = 1 | \textbf{Y}^{k-1}_0)$.  Here, we will assume that $\textbf{Z}_k = \textbf{0}$ is the unique absorbing state of the system.   

The Bayes update for $p_{i,k}$ can be written as ($i$ subscript omitted for clarity):
\begin{eqnarray}\label{bayesupdating}
	p_k & =& \frac{  f^{(1)}_k  }{  f^{(1)}_k p_{k|{k-1}} +  f^{(0)}_k (1-p_{k|{k-1}} )} p_{k|{k-1}} \nonumber \\
	& =& \frac{  f^{(1)}_k / f^{(0)}_k   }{1 +  \frac{   f^{(1)}_k - f^{(0)}_k  }{  f^{(0)}_k} p_{k|{k-1}}  } p_{k|{k-1}} \nonumber \\
	& =& \frac{  f^{(1)}_k / f^{(0)}_k   }{1 +  \frac{ \Delta f_k }{ f^{(0)}_k } p_{k|{k-1}}  }  p_{k|{k-1}}\label{bayes_geo}.
\end{eqnarray}
There are three different sampling/observation dependent possibilities for each individual at time $k$: case (1), $i$ is not sampled and therefore, $p_k = p_{k|k-1}$, case (2),  $\Delta f_k > 0$, and case (3),  $\Delta f_k < 0$.

We first derive a tight-upper bound for cases (2) and (3) of the form $p_k \le c_k \ p_{k|k-1}$.  For the remainder of the analysis we will assume that $| \frac{ \Delta f_k }{ f^{(0)}_k } p_{k|{k-1}} | < 1$ for cases (2) and (3) (see Appendix).

Using the upper-bounds derived in the Appendix, and after gathering all $n$ nodes, we have the following element-wise upper-bound on the updated belief state:
\begin{equation}\label{upper_pos}
	\textbf{p}_k  \le  \textbf{C}_k \textbf{p}_{k|{k-1}} =  \left( \textbf{B}_k  + \mathcal{O}_k \right) \textbf{p}_{k|{k-1}}.           
\end{equation}
with $ \textbf{B}_k = \text{diag}\left(b_{i,k}\right)$ and $ \mathcal{O}_k  = \text{diag}\left(    \mathbb{I}_{ \{ \Delta f_{i,k} < 0 \} }      \mathcal{O}\left(  \frac{|\Delta f_{i,k}|   }{ f^{(0)}_{i,k} }  p_{i,k|{k-1}} \right) \right)$ where $\mathbb{I}_{ \{ \Delta f_{i,k} < 0 \} }$ is the indicator function for the event $\Delta f_{i,k} < 0$.

Thus far, we have established, under the assumptions of $| \frac{ \Delta f_k }{ f^{(0)}_k } p_{k|{k-1}} | < 1$, an upper-bound for the updated posterior in terms of observation likelihoods and the belief state (\ref{upper_pos}).

Next, consider the restricted class of two-state Markov processes on $\mathcal{G}$, for which we can produce a bound of the form
\begin{equation}\label{dynamic_bound}
	\textbf{p}_{k|k-1} \le \textbf{S} \textbf{p}_{k-1}
\end{equation}
where \textbf{S} contains information about the transition parameters and the topology of the network.

It turns out that the $SIS$ model of mathematical epidemiology falls within this restricted class of percolation problems \cite{chak08}.

\noindent
The $SIS$ model on a graph $\mathcal{G}$, assumes that each of the $n$ individuals are in states $0$ or $1$, where $0$ corresponds to \textit{susceptible} and $1$ corresponds to \textit{infected}.  At any time $k$, an individual can receive the infection from their neighbors, $\eta(i)$, based upon their states at $k-1$.

\begin{figure}[h!]
\centering
\includegraphics[width=1.80in]{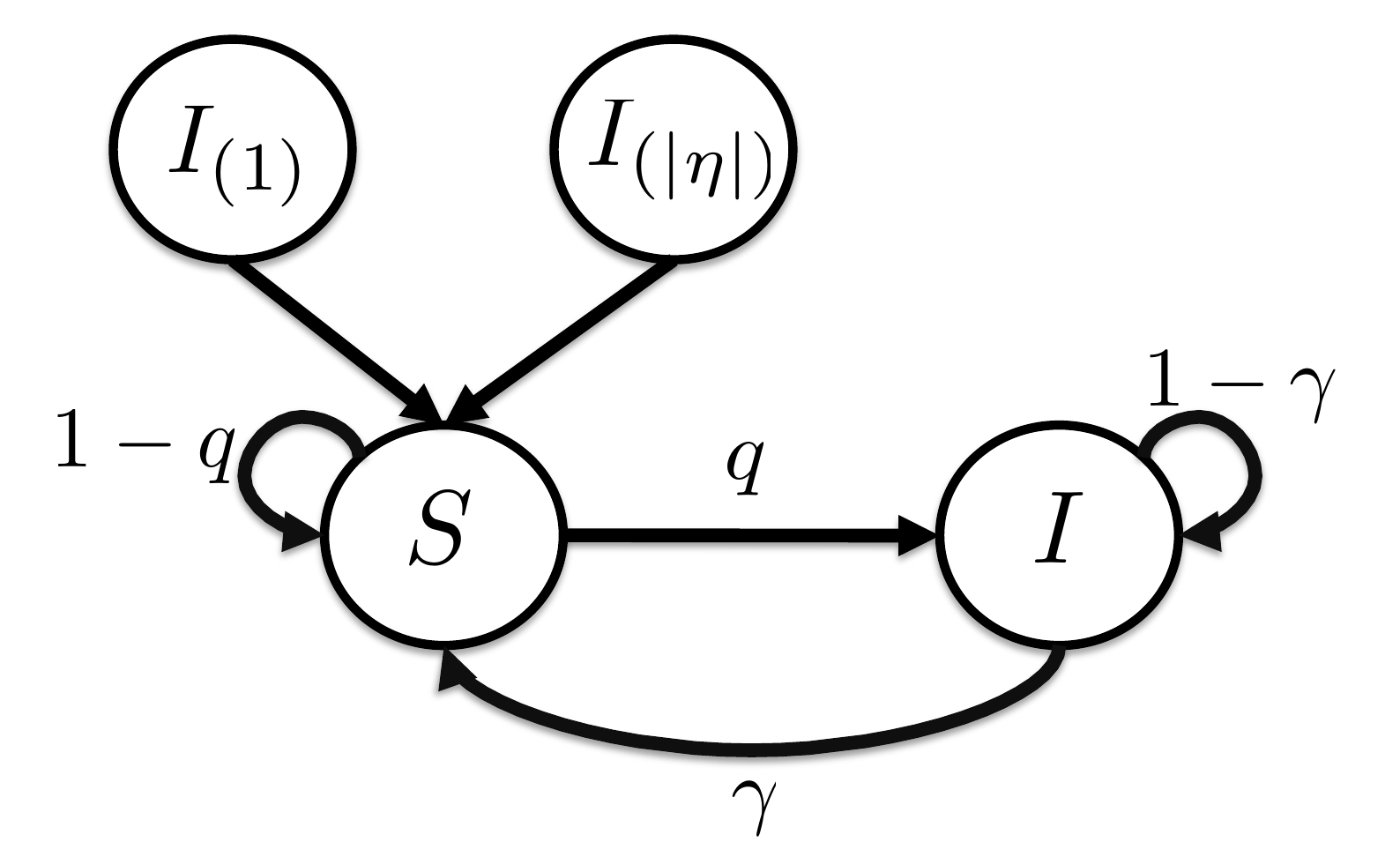}
\caption{$SIS$ Markov Chain for Node $i$ Interacting with the Infected States of its Neighbors}
\label{SIS_chain}
\end{figure}

\noindent
Under this $SIS$ model in \cite{chak08}
\begin{equation}\label{system_matrix}
	\textbf{S} = (1-\gamma)\textbf{I} + \beta \textbf{A}
\end{equation}	
where the Markov transition parameters $\gamma$ is the probability of $i$ transitioning from $1$ to $0$, $\beta$ is the probability of transmission between neighbors $i$ and $j$, and \textbf{A} is the graph adjacency matrix (see Figure \ref{SIS_chain}).  

Returning to the derivation, using the bound (\ref{dynamic_bound}), we have, by induction, the following recursion:
\begin{eqnarray}\label{recursion}
	\textbf{p}_k  &\le&  \textbf{C}_k \textbf{p}_{k|{k-1}} \le  \textbf{C}_k \textbf{S} \textbf{p}_{k-1} \le  \left(  \textbf{C}_k \textbf{S} \cdots  \textbf{C}_1 \textbf{S} \right) \textbf{p}_0 \nonumber \\
	 &=& \left( \textbf{B}_k \textbf{S} \cdots \textbf{B}_1 \textbf{S} \right) \textbf{p}_0 + \mathcal{O}_{  \textbf{C}_k \textbf{S} }      
\end{eqnarray}
where we have lumped the higher order modes and higher order cross-terms into $\mathcal{O}_{ \textbf{C}_k \textbf{S} } $.

The \textit{dominant mode of decay} of the updated posterior may be found by investigating the following eigen-decomposition:
\begin{equation}\label{BS}
	\textbf{B}_k \textbf{S} = \left(  \sum_{j=1}^n b_{j,k} \textbf{e}_j \textbf{e}_j^T   \right)  \left(  \sum_{j=1}^n \lambda_j \textbf{u}_j \textbf{u}_j^T   \right)
\end{equation} 
with $\textbf{e}_j = [0,\dots,0,1,0,\dots,0]^T$ ($1$ at $j^{th}$ element).  Without loss of generality, we can assume the eigenvalues of $\textbf{S}$ are listed in decreasing order, $|\lambda_1| \ge \dots \ge |\lambda_n|$.  Now rewriting (\ref{BS}), we have
\begin{eqnarray}\label{spectral}
	\textbf{B}_k \textbf{S} &=& \left( b_{j_k} \textbf{e}_{j_k} \textbf{e}_{j_k}^T  + \mathcal{O}_B   \right)  \left( \lambda_1 \textbf{u}_1 \textbf{u}_1^T +  \mathcal{O}_S   \right) \nonumber \\
	 &=& \left(  \lambda_1 b_{j_k} \textbf{e}_{j_k} \textbf{e}_{j_k}^T  \textbf{u}_1 \textbf{u}_1^T + \mathcal{O}_{BS}       \right)     
\end{eqnarray}
where $b_{j_k} = \text{max}_{j\in\{1,\dots,n\}} b_{j,k}$ and the $\mathcal{O}_B, \mathcal{O}_S, \mathcal{O}_{BS}$ variables corresponds to the higher order terms.  Inserting (\ref{spectral}) into (\ref{recursion}), and matching the largest eigenvalues of $\textbf{B}_k$ with $\lambda_1$ we obtain
\begin{eqnarray}\label{bound}
	\textbf{p}_k &\le& \left( \textbf{B}_k \textbf{S} \dots \textbf{B}_1 \textbf{S} \right) \textbf{p}_0 + \mathcal{O}_{ \textbf{C}_k \textbf{S} } \nonumber \\
	 &=& \hspace{-3mm}  \lambda_1^k \prod_{l=1}^k b_{j_l} \left( \prod_{l=1}^k\left( \textbf{e}_{j_l} \textbf{e}_{j_l}^T  \textbf{u}_1 \textbf{u}_1^T \right) \right) \textbf{p}_0 + \hspace{-1mm} \mathcal{O}(\varphi^k).
\end{eqnarray}
Thus, at large $k$, the dominant mode of the posterior goes as $\lambda_1^k \prod_{l=1}^k b_{j_l}$ (the modes in $\mathcal{O}(\varphi^k)$ decay faster than the dominant mode presented above).  

We can see that if the spectral radius of $\textbf{S}$ is less than one, $|\lambda_1| < 1$, then for large $k$, $\textbf{p}_k \to \textbf{0}$, which is the unique absorbing state of the system.    

This epidemic threshold condition on $\lambda_1$ has been previously established for unforced $SIS$-percolation processes \cite{chak08}.  However, in the tracking framework, the rate at which the posterior decays to the \textit{susceptible} state is perturbed by an additional measurement dependent factor, $\prod_{l=1}^k b_{j_l}$.  

This measurement-dependent dominant mode of the posterior should more accurately model the true dynamic response of the node states better than that in \cite{chak08} since the posterior better tracks the truth than the unforced predictive distribution.  Additionally, this dominant mode of the updated posterior distribution allows one to simulate the response of the percolation threshold to intervention and control actions which are designed to increase the threshold, such that the probability of epidemics is minimized.




\section{NUMERICAL EXAMPLE}

\begin{figure}[h!] 
\vspace{-3.0mm} 
\centering
\includegraphics[width=2.5in]{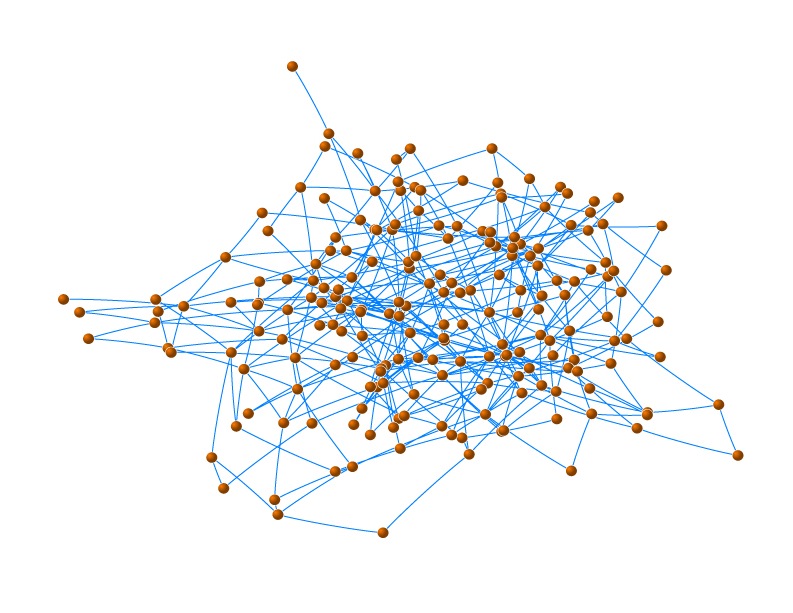}
\vspace{-2mm}
\caption{200 Node Scale-Free Graph $\mathcal{G} = (\mathcal{V},\mathcal{E})$} 
\label{sf}
\end{figure}

Here, we present results of simulations of our adaptive sampling for the active tracking of a causal Markov ground truth process across a random 200 node, scale-free network (Figure \ref{sf}).  Since the goal in tracking is to accurately classify the states of each node, we are interested in exploring the detection performance as the likelihood of an epidemic increases through the percolation threshold for this graph.  

One would expect different phase transitions (thresholds) in detection performance for various sampling strategies, ranging from the lowest threshold for unforced percolation distributions to highest for a continuous monitoring of all $n$ nodes. We will present a few of these detection surfaces that depict these phase transitions for the unforced percolation distribution, random $m=40$ node sampling, and our proposed information theoretic adaptive sampling of $m=40$.

Here, we will restrict our simulations to the two-state $SIS$ model of mathematical epidemiology described above.

The sensor models (\ref{Y}), are of the form of two-dimensional multivariate Guassians with common covariance and shifted mean vector.  The transition dynamics of the $i^{th}$ individual (\ref{Z}), for the $SIS$ model is given by:

\vspace{-6mm}
\begin{small}
\begin{equation}\label{SIS}
	Z^i_k|\textbf{Z}^{ \{i,\eta(i)\}  }_{k-1}  \hspace{-1mm} \sim (1-\gamma) Z^i_{k-1} + (1 - Z^i_{k-1}) \hspace{-1mm} \left[ 1- \hspace{-2mm}  \displaystyle\prod_{j \in \eta(i)} ( 1-\beta  Z^j_{k-1}   )\right].
\end{equation}
\end{small}
\vspace{-4mm}

where $Z^i_{k-1} \in \{0,1\}$ is the indicator function of $i$ being infected at time $k-1$.  The transmission term between $i$ and $\eta(i)$ is known the Reed-Frost model \cite{chak08,draief-2006,newman-2002-66}.  Since the tail of the degree distribution of our synthetic scale-free graph contains nodes with degree greater than 10, updating (\ref{marginal}) exactly is unrealistic and we must resort to approximate algorithms.  Here, we will assume the mean field approximation used by \cite{chak08} for this $SIS$ model, resulting in the following marginal belief state update for the $i^{th}$ node of infected ($Z_k^i = 1$):     

\vspace{-6mm}
\begin{small} 
\begin{equation}\label{mfapprox}
	p_{i,k|k-1} = (1-\gamma)p_{i,k-1} + (1 -  p_{i,k-1}) \left[ 1-\displaystyle\prod_{j \in \eta} ( 1-\beta p_{j,k-1}  )\right].
\end{equation}
\end{small}
\vspace{-4mm}

Equation (\ref{mfapprox}) allows us to efficiently update the marginal belief state directly for all $n$ nodes which are then used for estimating the best $m$ measurements using (\ref{IG_marg}).
\begin{figure}[h!]
\centering
\subfigure[AUR Surface for Unforced Prediction Distribution (no evidence acquired throughout the monitoring)] 
{
    \label{AUR_perc}
    \includegraphics[width=8.5cm]{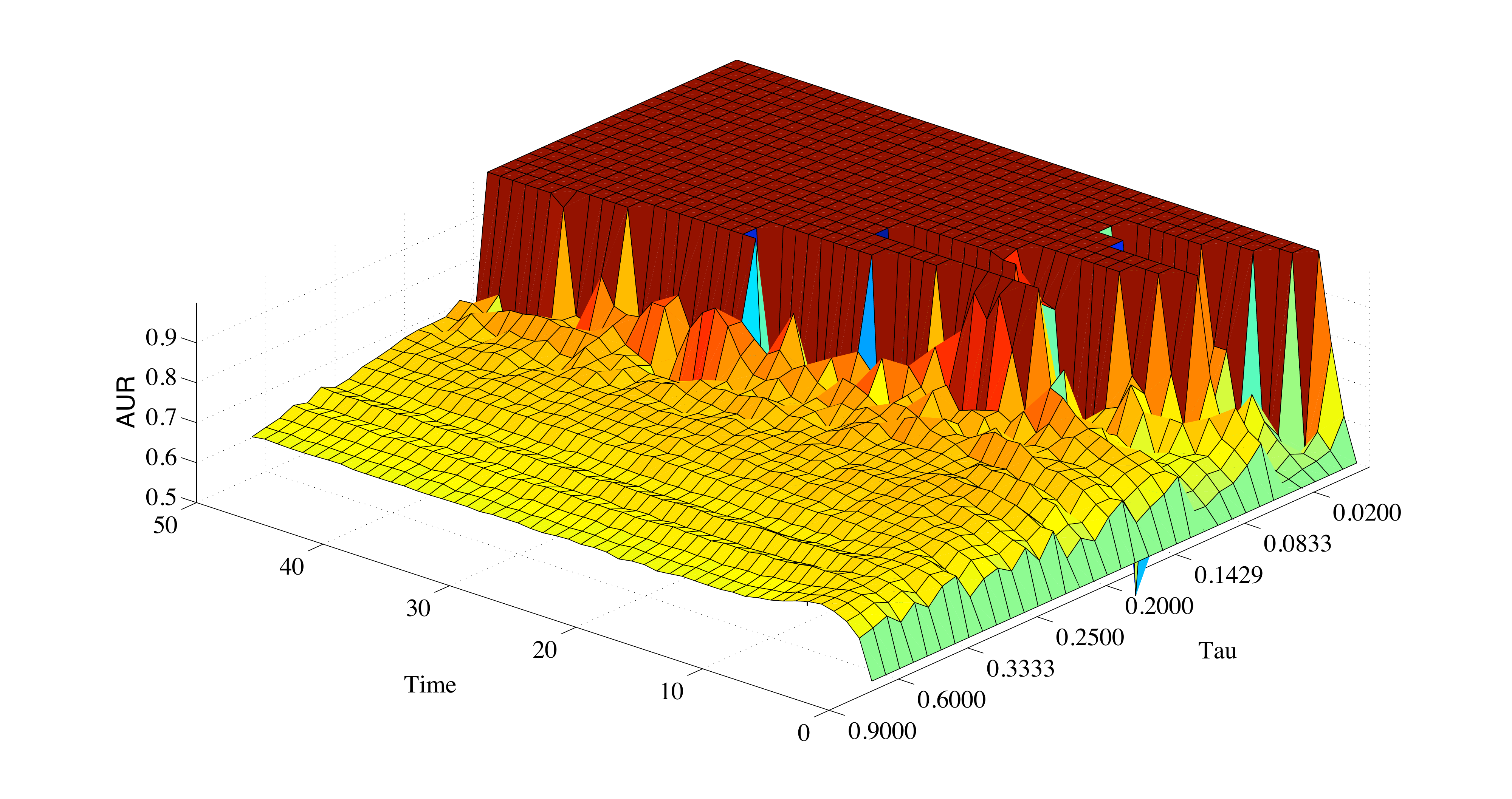}
}
\hspace{1cm}
\subfigure[AUR Surface for Updated Posterior Distribution with $m=40$ Random Measurements at Each Time $k$] 
{
    \label{AUR_rnd}
    \includegraphics[width=8.5cm]{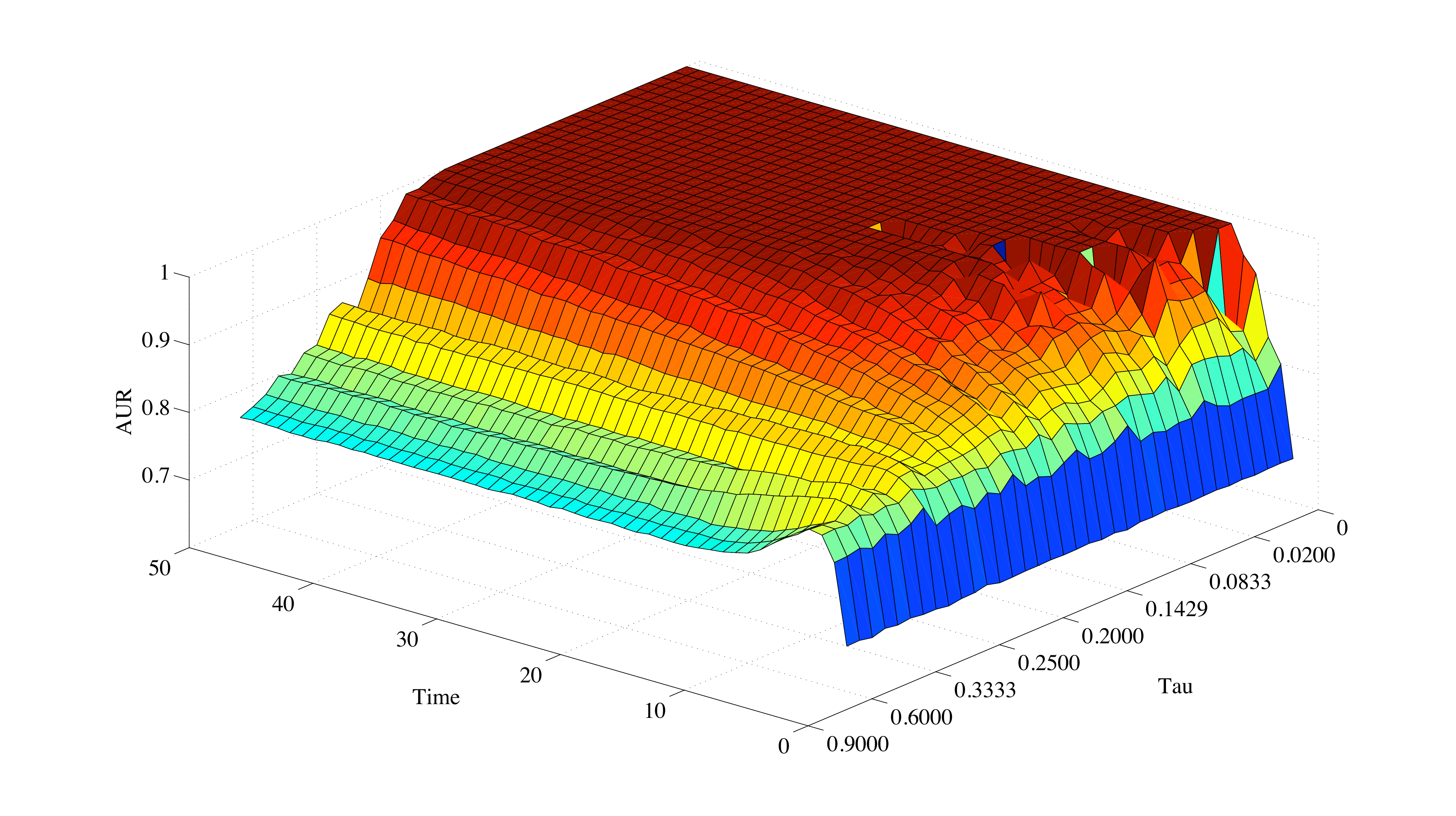}
}
\hspace{1cm}
\subfigure[AUR Surface for Updated Posterior Distribution with $m=40$ Information Theoretic Adaptive Measurements at Each Time $k$] 
{
    \label{AUR_adaptive}
    \includegraphics[width=8.5cm]{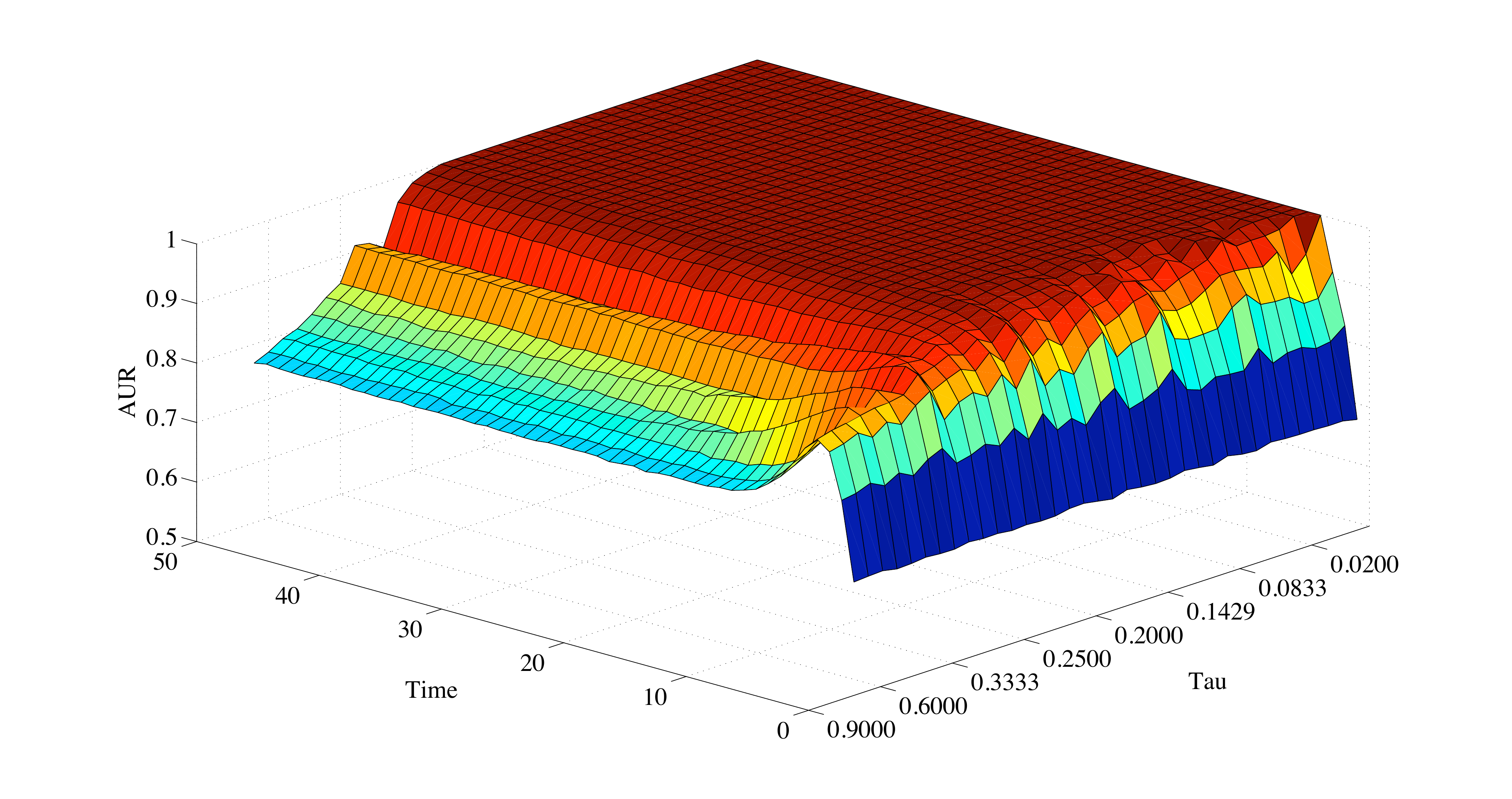}
}
\caption{Area under the ROC curve surface as a function of percolation parameter $\tau = \beta/\gamma$ and time }
\label{AUR} 
\end{figure}

As we are interested in detection performance as a function of time and epidemic intensity, the Area Under the ROC Curve (AUR) is a natural statistic to quantify the detection power (detection of the infected state).  The AUR is evaluated at each time $k$, each $SIS$ percolation intensity parameter
\begin{equation}\label{tau}
 	\tau = \beta/\gamma
\end{equation}
 and over 500 random initial states of the network.  For the $SIS$ model, $\tau$ is the single parameter (aside from the topology of the graph) that characterizes the intensity of the percolation/epidemic.  It is useful to understand how the detection performance varies as a function of epidemic intensity, as it indicates how well the updated posteriors are playing ``catch-up'' in tracking the true dynamics on the network.   

For this $SIS$ model, the percolation threshold is defined as $\tau_c = 1/\lambda_1(\textbf{A})$ where $\lambda_1(\textbf{A}) = \text{max}_{i \in \{1,\dots,n\} } | \lambda_i |$ is the spectral radius of the graph adjacency matrix, $\textbf{A}$ \cite{chak08}.  Values of $\tau$ greater than $\tau_c$ imply that any infection tend to become an epidemic, whereas those values less than $\tau_c$ imply that small epidemics tend to die out.

For the network under investigation (Figure \ref{sf}), $\tau_c = 0.1819$.  We see from Figure \ref{AUR_perc} that a phase transition in detection power (AUR) for the unforced percolation distribution does indeed coincide with the epidemic threshold $\tau_c$.  While the epidemic threshold for the random and adaptive sampling policies is still $\tau_c = 0.1819$, the measurements acquired allow the posterior to better track the truth, but only up to their respective phase transitions in detection power (see Figures \ref{AUR_rnd} and \ref{AUR_adaptive}).    

Figure \ref{AUR_adaptive} confirms that the adaptive sampling better tracks the truth than randomly sampling nodes, while pushing the phase transition in detection performance to higher percolation intensities, $\tau$.  We see that the major benefit of the adaptive sampling is apparent when conditions of the network are changing moderately, at medium epidemic conditions.  Beyond a certain level of percolation intensity, more resources will need to be allocated to sampling to maintain a high level of detection performance.

A heuristic sampling strategy based on the topology of $\mathcal{G}$ was also explored (results not shown) by sampling the "hubs" (highly-connected nodes).  However, detection performance was only slightly better than random sampling and poorer than our adaptive sampling method.  

\begin{figure}[h!]
\centering
\includegraphics[width=3.60in]{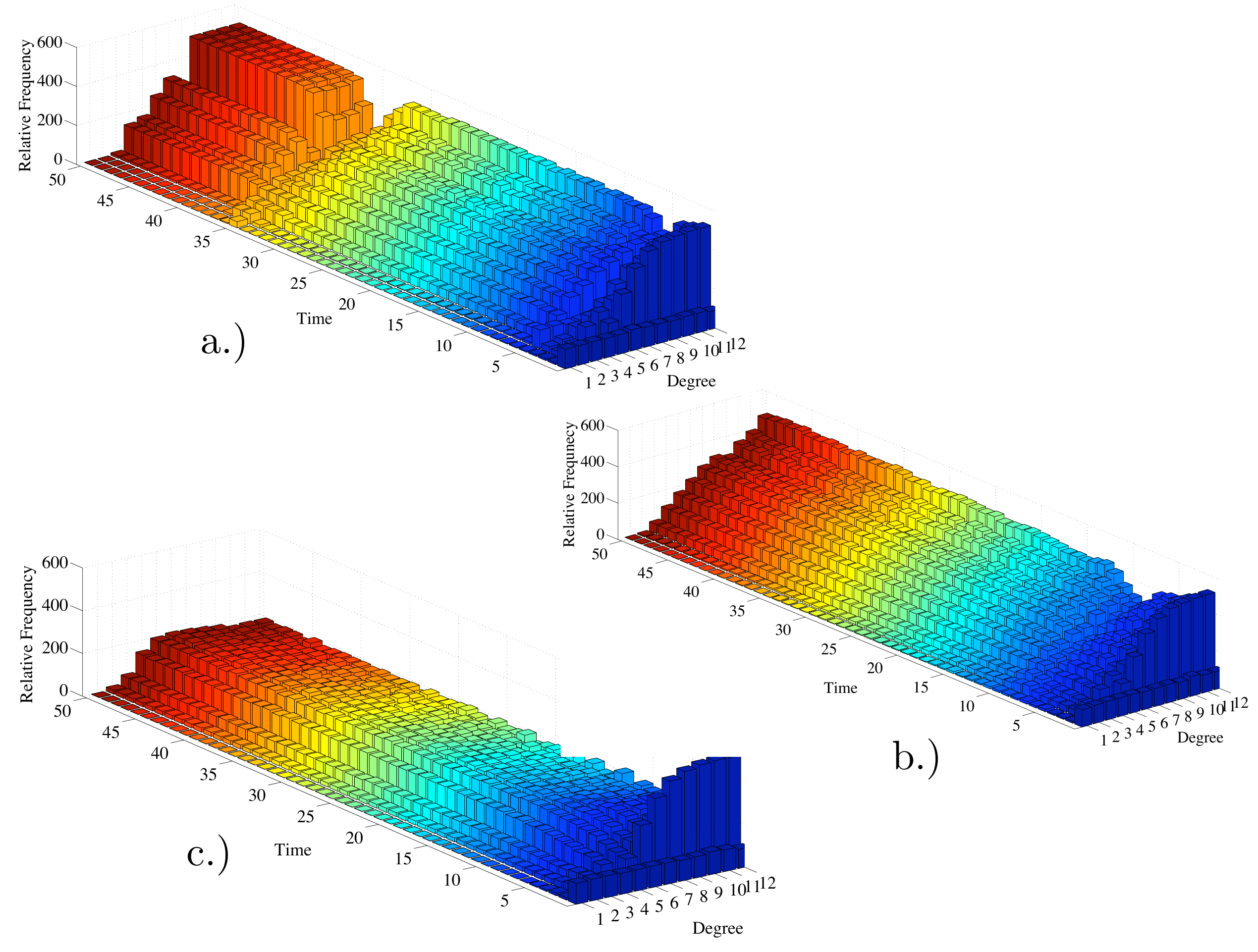}
\vspace{-10mm}
\caption{Relative Frequency of Nodes Sampled ($z$-axis) of a Given Degree ($x$-axis) Over Time ($y$-axis) for $m=40$ Adaptive Sampling Strategy: a.) $\tau = 0.125$, b.) $\tau = 0.2143$, and c.) $\tau = 0.5$}
\label{sampling}
\end{figure}

It is often useful for developing sampling heuristics and offline control/intervention policies to inspect what \textit{type} of nodes, topologically speaking, is the adaptive sampling strategy targeting, under various network conditions (different values of $\tau$).  In Figure \ref{sampling}, the relative frequency of nodes sampled with a particular degree is plotted against time (under the $m=40$ adaptive sampling strategy) for three different values of $\tau$ (over 500 random initial conditions of the network). 

For the larger of the three values explored ($\tau = 0.5 > \tau_c$) we see that the sampling is approximately uniform across the nodes of each degree on the graph (Figure \ref{sampling}(c)).  Therefore, under extremely intense epidemic conditions, the adaptive sampling strategy is targeting all nodes of each degree equally, and therefore, it is sufficient to perform random sampling.  For the two lower values of $\tau$, Figure \ref{sampling}(a) and Figure \ref{sampling}(b) (near $\tau_c$), we see that adaptive policy targets highly connected nodes more frequently than those of lesser degree and thus, it is more advantageous to exploit such a strategy, as compared to random sampling (see AUR surface in Figure \ref{AUR_adaptive}).

\section{DISCUSSION}
In this paper, we have derived the conditions for a network specific percolation threshold using expressions for the updated posterior distribution resulting from actively tracking the process.  These conditions recover the unforced percolation threshold derived in \cite{chak08} but with an additional factor involving sensor likelihood terms due to measurements obtained throughout the monitoring.  A term of the form $\lambda_1^k \prod_{l=1}^k b_{j_l}$ (derived in (\ref{bound})) was shown to be the dominant mode of the updated posterior dynamic response to active intervention of immunizing the nodes (holding node states constant).  The conditions of the percolation using the updated posterior should more accurately model the phase transition corresponding to a particular disease trajectory and therefore,  enable a better assessment of immunization strategies and any subsequent observations resulting from such actions.  The framework presented above, along with the new posterior percolation threshold, should provide additional insight into active monitoring of large complex networks under resource constraints.

\section{APPENDIX}\label{A}
In case (2), when $\Delta f_k > 0$, we can re-write (\ref{bayes_geo}) in terms of an \textit{alternating geometric series}:    
\begin{eqnarray}\label{bayesupdate_withsum}
	p_k &=& \frac{ f^{(1)}_k  }{  f^{(0)}_k  } \left[  \displaystyle\sum_{l=0}^{\infty} (-1)^l \left(\frac{  |\Delta f_k|   }{ f^{(0)}_k }  p_{k|{k-1}} \right)^l    \right]               p_{k|{k-1}} \nonumber \\
	&\le& \frac{ f^{(1)}_k  }{  f^{(0)}_k  }   \left[ 1 +  \frac{ |\Delta f_k|   }{ f^{(0)}_k }  p_{k|{k-1}}     \right]  p_{k|{k-1}}          
\end{eqnarray}
where we have used the fact that $1/(1+|a|) \le 1+|a|$.  Recalling that $p \ge p^2$ for $0 \le p \le 1$, we have
\begin{equation}\label{bayesupdate_withsum}
	p_k  \le \frac{ f^{(1)}_k  }{  f^{(0)}_k  }   \left[ 1 +  \frac{ |\Delta f_k|   }{ f^{(0)}_k }  \right] p_{k|{k-1}}.           
\end{equation}

In case (3), when $\Delta f_k < 0$, and (\ref{bayes_geo}) can be represented as a \textit{geometric series}:

\vspace{-5mm}
\begin{small}
\begin{gather}\label{bayesupdate_withsum}
	p_k = \frac{ f^{(1)}_k  }{  f^{(0)}_k  } \left[  \sum_{l=0}^{\infty} \left( \frac{  |\Delta f_k|   }{ f^{(0)}_k }  p_{k|{k-1}} \right)^l    \right]  p_{k|{k-1}} \nonumber \\
	= \frac{ f^{(1)}_k  }{  f^{(0)}_k  } \left[ 1 + \frac{  |\Delta f_k|   }{ f^{(0)}_k }  p_{k|{k-1}} + \sum_{l=2}^{\infty} \left(\frac{  |\Delta f_k|   }{ f^{(0)}_k }  p_{k|{k-1}} \right)^l    \right]       p_{k|{k-1}}.        
\end{gather}
\end{small}
\vspace{-5mm}

Once again, using the $p \ge p^2$ bound, we obtain:
\begin{equation}\label{bayesupdate_withsum}
	p_k  \le \frac{ f^{(1)}_k  }{  f^{(0)}_k  }   \left[ 1 +  \frac{ |\Delta f_k|   }{ f^{(0)}_k }  \right] p_{k|{k-1}} + \mathcal{O}\left(  \frac{|\Delta f_k|   }{ f^{(0)}_k }  p_{k|{k-1}} \right).        
\end{equation} 

A general inequality, that is equality for case (1), is of the form $p_k \le c_k \ p_{k|k-1}$ with
\begin{center} 
$b_k=\begin{cases}
1 &, i \notin \textbf{a}_k \\
\frac{ f^{(1)}_k  }{  f^{(0)}_k  }   \left[ 1 +  \frac{ |\Delta f_k|   }{ f^{(0)}_k }  \right] &, \Delta f_k > 0 \ \text{or} \ \Delta f_k < 0  
\end{cases}$
\end{center}
with $c_k  = b_k$ for cases (1) and (2) and $c_k  = b_k + \mathcal{O}\left(  \frac{|\Delta f_k|   }{ f^{(0)}_k }  p_{k|{k-1}} \right)$ for case (3).


\begin{thebibliography}{}

\end{thebibliography}


\begin{thebibliography}{10}

\bibitem{chak08}
{\sc D.~Chakrabarti, Y.~Wang, C.~Wang, J.~Leskovec, and C.~Faloutsos}, {\em
  Epidemic thresholds in real networks}, ACM Trans. Inf. Syst. Secur., 10
  (2008), pp.~1094--9224.

\bibitem{cohen-2003-91}
{\sc R.~Cohen, S.~Havlin, and D.~ben Avraham}, {\em Efficient immunization
  strategies for computer networks and populations}, Physical Review Letters,
  91 (2003), p.~247901.

\bibitem{Doucet:2000bv}
{\sc A.~Doucet, S.~Godsill, and C.~Andrieu}, {\em On sequential monte carlo
  sampling methods for bayesian filtering}, Statistics and Computing, 10
  (2000), pp.~197--208.

\bibitem{draief-2006}
{\sc M.~Draief, A.~J. Ganesh, and L.~Massouli{\'e}}, {\em Thresholds for virus
  spread on networks}, Ann. Appl. Probab., 18 (2008), pp.~359--378.

\bibitem{Eubank_2004}
{\sc S.~Eubank, H.~Guclu, Anil, M.~V. Marathe, A.~Srinivasan, Z.~Toroczkai, and
  N.~Wang}, {\em Modelling disease outbreaks in realistic urban social
  networks}, Nature, 429 (2004), pp.~180--184.

\bibitem{hero_book_07}
{\sc A.~O. Hero, D.~Castenon, D.~Cochran, and K.~D. Kastella}, {\em Foundations
  and applications of sensor management}, Springer, NY, 2007.

\bibitem{Mackay_2002_information}
{\sc D.~J.~C. Mackay}, {\em Information Theory, Inference \& Learning
  Algorithms}, Cambridge University Press, 2002.

\bibitem{newman-2002-66}
{\sc M.~E.~J. Newman}, {\em The spread of epidemic disease on networks},
  Physical Review E, 66 (2002), p.~016128.

\bibitem{newman-2005-95}
{\sc M.~E.~J. Newman}, {\em Threshold effects for two pathogens spreading on a
  network}, Physical Review Letters, 95 (2005), p.~108701.

\bibitem{Ng02factoredparticles}
{\sc B.~Ng, L.~Peshkin, and A.~Pfeffer}, {\em Factored particles for scalable
  monitoring}, in In Proceedings of the Eighteenth Conference on Uncertainty in
  Artificial Intelligence, 2002, pp.~370--377.

\bibitem{ecs16801}
{\sc Z.~Rabinovich and J.~S. Rosenschein}, {\em Extended markov tracking with
  an application to control}, in The Third International Joint Conference on
  Autonomous Agents and Multiagent Systems, 2004, pp.~95--100.

\bibitem{Zheng05}
{\sc A.~X. Zheng, I.~Rish, and A.~Beygelzimer}, {\em Efficient test selection
  in active diagnosis via entropy approximation}, in Proceedings of the 21th
  Annual Conference on Uncertainty in Artificial Intelligence, 2005, p.~675.

\end{thebibliography}
\end{document}